%% file: main.tex
\documentclass[10pt,twocolumn,letterpaper]{article}

\usepackage{cvpr}              
\input{preamble}

\definecolor{cvprblue}{rgb}{0.21,0.49,0.74}
\usepackage[pagebackref,breaklinks,colorlinks,allcolors=cvprblue]{hyperref}

\def\paperID{7660}
\def\confName{CVPR}
\def\confYear{2025}

\title{\vqlb{}: A Simple Training-Free Baseline for Visual Query Localization Using Region-Based Representations}

\author{
    Savya Khosla \quad Sethuraman T V \quad Alexander Schwing \quad Derek Hoiem \\
    \small University of Illinois Urbana-Champaign \\
    \tt\small \{savyak2, st34, aschwing, dhoiem\}@illinois.edu
}

\begin{document}
\maketitle

\begin{figure*}[t]
    \centering
    \includegraphics[width=0.95\linewidth]{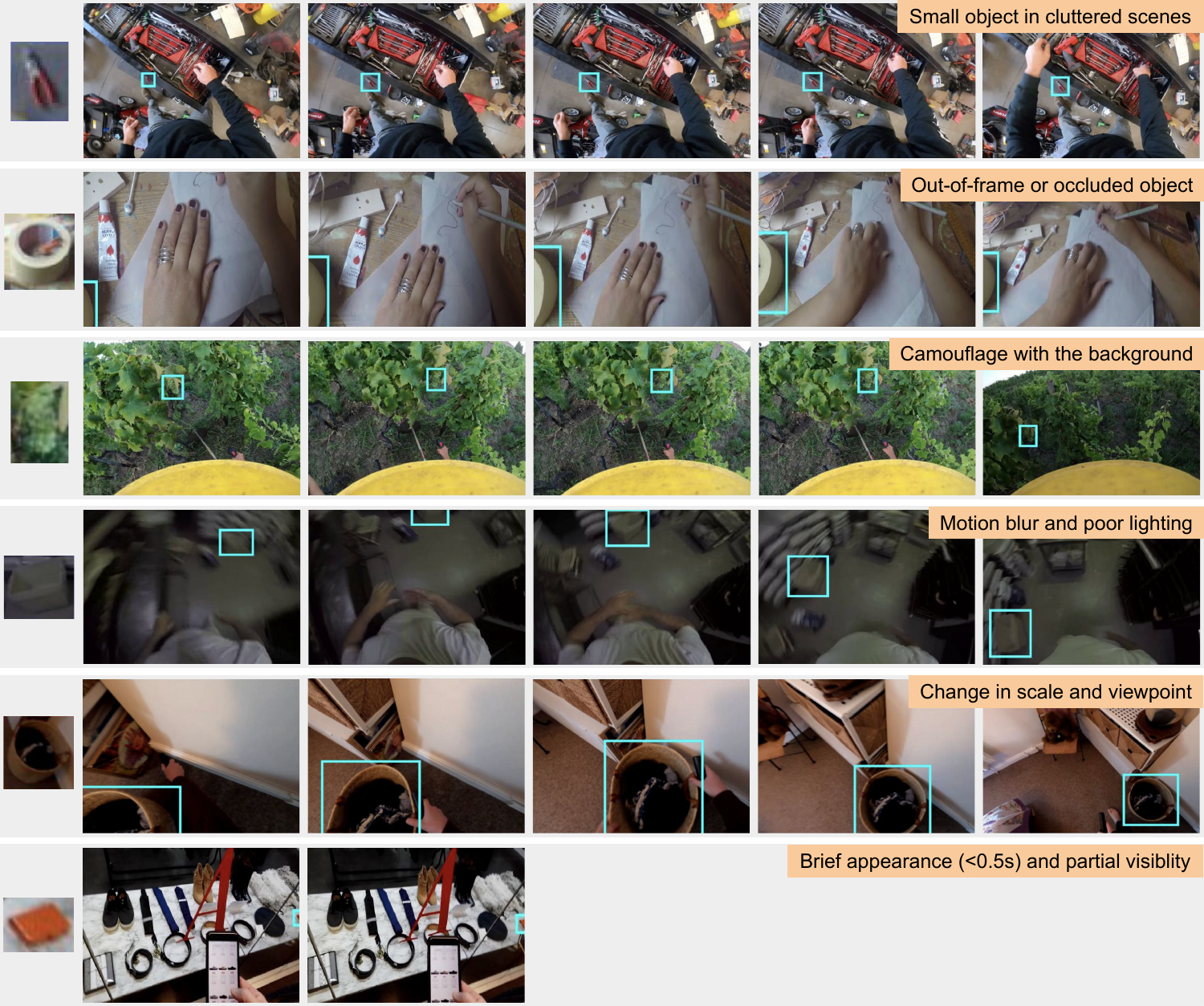}
    \caption{\vql{} is a training-free framework designed for visual query localization. It can effectively localize target objects in long videos despite challenging conditions such as visual clutter, occlusions, background blending, motion blur, viewpoint changes, and brief object appearances. Here we show visual queries on the left and the successfully localized object appearances marked by cyan bounding boxes on the right.}
    \label{fig:teaser}
\end{figure*}

\input{sections/0-abstract}    
\input{sections/1-intro}
\input{sections/2-related}
\input{sections/3-method}
\input{sections/4-experiments}
\input{sections/5-conclusion}

\section*{Acknowledgements}
This work is supported in part by ONR N00014-23-1-2383, DARPA HR0011-23-9-0060, NSF 2008387, NSF 2045586, NSF 2106825, NSF MRI 1725729, and NIFA award 2020-67021-32799. We used GPUs at NCSA Delta through allocation CIS240541 from the Advanced Cyberinfrastructure Coordination Ecosystem: Services \& Support (ACCESS) program, which is supported by U.S. National Science Foundation grants \#2138259, \#2138286, \#2138307, \#2137603, and \#2138296. The views and conclusions expressed are those of the authors, and not necessarily representative of the US Government or its agencies.

{
    \small
    \bibliographystyle{ieeenat_fullname}
    \bibliography{main}
}

\input{sections/6-supplementary}

\end{document}

%% file: preamble.tex
\newcommand{\vql}{\text{\sc{Relocate}}}
\newcommand{\vqlb}{\mbox{\sc{\textbf{Relocate}}}}
\newcommand{\boldheader}[1]{\noindent\textbf{#1.}}

\usepackage[skip=10pt]{caption}
\usepackage{graphicx}
\usepackage{subcaption}
\usepackage{multirow}
\usepackage{booktabs}
\usepackage{colortbl}
\usepackage{xcolor}
\definecolor{lightblue}{rgb}{0.4, 0.6, 1}
\definecolor{lightgreen}{rgb}{0.3, 0.8, 0.4}
\definecolor{lightgreen}{rgb}{0.3, 0.8, 0.4}
\definecolor{lightred}{rgb}{1.0, 0.4, 0.4}
\definecolor{lightorange}{rgb}{0.95, 0.65, 0.4}
\definecolor{mauve}{rgb}{0.8, 0.6, 0.9}

%% file: sections/0-abstract.tex
\begin{abstract}
We present \vql{}, a simple training-free baseline designed to perform the challenging task of visual query localization in long videos. To eliminate the need for task-specific training and efficiently handle long videos, \vql{} leverages a region-based representation derived from pretrained vision models. At a high level, it follows the classic object localization approach: (1) identify all objects in each video frame, (2) compare the objects with the given query and select the most similar ones, and (3) perform bidirectional tracking to get a spatio-temporal response. However, we propose some key enhancements to handle small objects, cluttered scenes, partial visibility, and varying appearances. Notably, we refine the selected objects for accurate localization and generate additional visual queries to capture visual variations. We evaluate \vql{} on the challenging Ego4D Visual Query 2D Localization dataset, establishing a new baseline that  outperforms prior task-specific methods by 49\% (relative improvement) in spatio-temporal average precision.
\end{abstract}

%% file: sections/1-intro.tex
\section{Introduction}
\label{sec:intro}
Visual Query Localization (VQL) requires localizing the last appearance of an object of interest in a long video. The object of interest is specified via a reference image, also known as the \textit{visual query}. Figure~\ref{fig:teaser} provides illustrative examples. VQL is an important task, \eg, for surveillance, legal investigations, wildlife monitoring, or simply for tracking down a misplaced item. However, the task presents several unique challenges that push the boundaries of contemporary computer vision methods. For instance, unlike classic object detection models that are trained to identify a fixed set of object categories~\cite{Ren2015FasterRT, Carion2020EndtoEndOD}, VQL requires localizing an open-ended range of objects. Additionally, while typical object tracking methods are initialized with a bounding box close to the object's temporal location in the video~\cite{Yan2021LearningST, Ma2022UnifiedTT, Mayer2022BeyondST}, the VQL reference image (visual query) often originates outside the video, \ie, there may be no exact or neighboring frame for reliable matching. Further, the object's appearance may vary significantly from the visual query due to changes in orientation, scale, context, lighting, motion blur, and occlusions. Compounding these issues, the object of interest usually appears briefly in a long, untrimmed video.

Classic VQL methods typically use a stage-wise pipeline~\cite{Xu2022NegativeFM, Xu2022WhereIM, Grauman2021Ego4DAT}: (1) identify all objects in each video frame, (2) compare these objects with the given query to select the most similar ones, and (3) perform bidirectional tracking to obtain a spatio-temporal response. While this pipeline is well-grounded in the object localization literature, it is effective only for large, consistently visible objects that closely match the visual query in short video clips. A more recent work~\cite{Jiang2023SingleStageVQ} has proposed an end-to-end framework that aims to understand the holistic relationship between a given query and the video, performing spatio-temporal localization in a single step. However, this approach requires extensive training on large, annotated datasets, which can be challenging to obtain. Additionally, since such a method learns a holistic video-query relationship, it must process the entire video for each query, even when multiple queries are to be localized in the same video.

To address these limitations, we propose \underline{Re}gion-based representations for \underline{Loc}alizing \underline{A}ny\underline{t}hing in \underline{E}pisodic memory (\vql{}), a simple \emph{training-free} baseline for VQL. Being training-free, \vql{} eliminates the need for extensive task-specific training and large annotated datasets. Moreover, it encodes a video independently of the query, allowing the same video encoding to be reused for multiple queries. This  makes the method more suitable for episodic memory tasks, as it enables us to perform the resource-intensive video encoding just once for multiple queries. 

\vql{} follows a classic stage-wise setup for object localization while integrating techniques to efficiently encode long videos, handle small and fleeting objects, and manage varying appearances of the query object. Specifically, it begins by extracting region-based representations from the video and searching for candidates that match the visual query. The selected candidates are then refined to improve precision, and the most relevant candidate is tracked across video frames to generate an initial prediction. In a subsequent iteration, this prediction is used to create new visual queries, which are then employed to relocalize the object, aiming to capture appearances that differ significantly from the original visual query.

We evaluate \vql{} using the Ego4D Visual Query 2D (VQ2D) Localization benchmark~\cite{Grauman2021Ego4DAT} and observe significant improvements over prior methods. We achieve a 49\% increase in spatio-temporal average precision and a 33\% boost in temporal average precision compared to prior state-of-the-art, all without task-specific training.

In summary, the main contributions of this work are:

\begin{enumerate}
    \item We propose \vql{}, a \emph{training-free} method for localizing objects in long videos. Despite no task-specific training, \vql{} significantly improves upon prior VQL work on the  Ego4D VQ2D benchmark.
    \item We demonstrate the benefits of using region-based representations from pretrained vision models for object localization. Particularly, we show that these representations form a detailed yet compact encoding for long videos. Furthermore, they allow us to efficiently and robustly perform object retrieval using a simple matching function like cosine similarity.
    \item We propose techniques to improve the robustness and precision of a  stage-wise object localization framework. Specifically, to enhance localization accuracy, we introduce a refinement step that allows \vql{} to closely examine the selected candidates and improve their representations, while filtering out incorrect selections. Additionally, we introduce a technique to generate multiple visual queries from a single query, which helps capture varying appearances, changing contexts, and occlusions of objects in a dynamic video setting.
\end{enumerate}

%% file: sections/2-related.tex
\section{Related Work}
\label{sec:related-work}
\boldheader{Object Instance Recognition}
Early approaches to object instance recognition retrieve images with similar keypoints to the visual query~\cite{Sivic2003VideoGA, Nistr2006ScalableRW, Lowe2004DistinctiveIF, Obdrzlek2005SublinearIF}. This works well for objects with distinctive textures, but fails for low-texture, blurry, and highly occluded objects, which are common in the VQL task. Subsequent research shifts towards using features extracted from convolutional neural networks (CNNs) as descriptors for image retrieval~\cite{Babenko2014NeuralCF, Chandrasekhar2016APractical, Razavian2014CNNFO, Razavian2014ABF, Azizpour2014FromGT, Babenko2015AggregatingLD}. However, these early CNN-based methods suffer when faced with large changes in scale, rotation, or viewpoint.
The advent of vision foundation models such as CLIP~\cite{Radford2021LearningTV}, DINO~\cite{Caron2021EmergingPI}, and DINOv2~\cite{Oquab2023DINOv2LR} has enabled new strategies for instance recognition and retrieval. Several works~\cite{Baldrati2022EffectiveCA, Saito2023Pic2WordMP, Chen2020LearningJV, Sain2023CLIPFA} have leveraged the cross-modal capabilities of CLIP for specialized retrieval tasks. However, these approaches are primarily designed for scenarios where query objects occupy a substantial portion of the target frame, which is not the case in VQL. Recent work on region-based representations~\cite{ShlapentokhRothman2024RegionBasedRR} shows that pooling DINOv2 features over SAM regions provides effective example-based object category retrieval. In this work, we apply a similar approach for our initial search but propose enhancements to refine, track, and expand the initial query.

\boldheader{Video Object Tracking}
Conventional tracking approaches are initialized with a bounding box in an initial frame of the target video, and they track objects through gradual appearance changes. In contrast, VQL often involves queries from frames outside the target video, leading to significant variations in appearance, viewpoint, and context. Nevertheless, tracking remains crucial in VQL. Early tracking approaches relied on motion and appearance cues, using correlation filters and keypoint matching~\cite{Ross2008IncrementalLF, Henriques2014HighSpeedTW, Kalal2012TrackingLearningDetection}. More recently, trackers leverage large Transformer models to effectively track one or more objects in long videos~\cite{Chen2021TransformerT, Cui2022MixFormerET, Mayer2022TransformingMP, Ma2022UnifiedTT, Meinhardt2021TrackFormerMT, Yan2021LearningST}. Some recent works have also extended SAM~\cite{Kirillov2023SegmentA} for tracking~\cite{Cheng2023SegmentAT, Yang2023TrackAS, Ravi2024SAM2S}. In this work, we use SAM 2~\cite{Ravi2024SAM2S}.

\boldheader{Visual Query Localization} 
The Ego4D benchmark~\cite{Grauman2021Ego4DAT} recently introduced VQ2D, a benchmark for VQL. The initial approach, proposed as a baseline in Ego4D, employs a three-stage detection and tracking framework: performing frame-level detection, identifying the most recent detection peak across time, and applying bi-directional tracking to determine the complete temporal extent of the target object~\cite{Grauman2021Ego4DAT}. Subsequent works enhance the framework's performance through various refinements like incorporating negative frame sampling to reduce false positives~\cite{Xu2022NegativeFM} and leveraging background objects as contextual cues~\cite{Xu2022WhereIM}. In this work, we adopt a similar stage-wise framework, introducing key design decisions and refinements that significantly improve the baseline. A more recent effort trains a network to learn the query-video relationship and perform VQL in a single step. However, it relies on a large amount of annotated data, which is costly to obtain. In contrast, we present a training-free solution.

%% file: sections/3-method.tex
\begin{figure*}[t]
    \centering
    \includegraphics[width=0.98\linewidth]{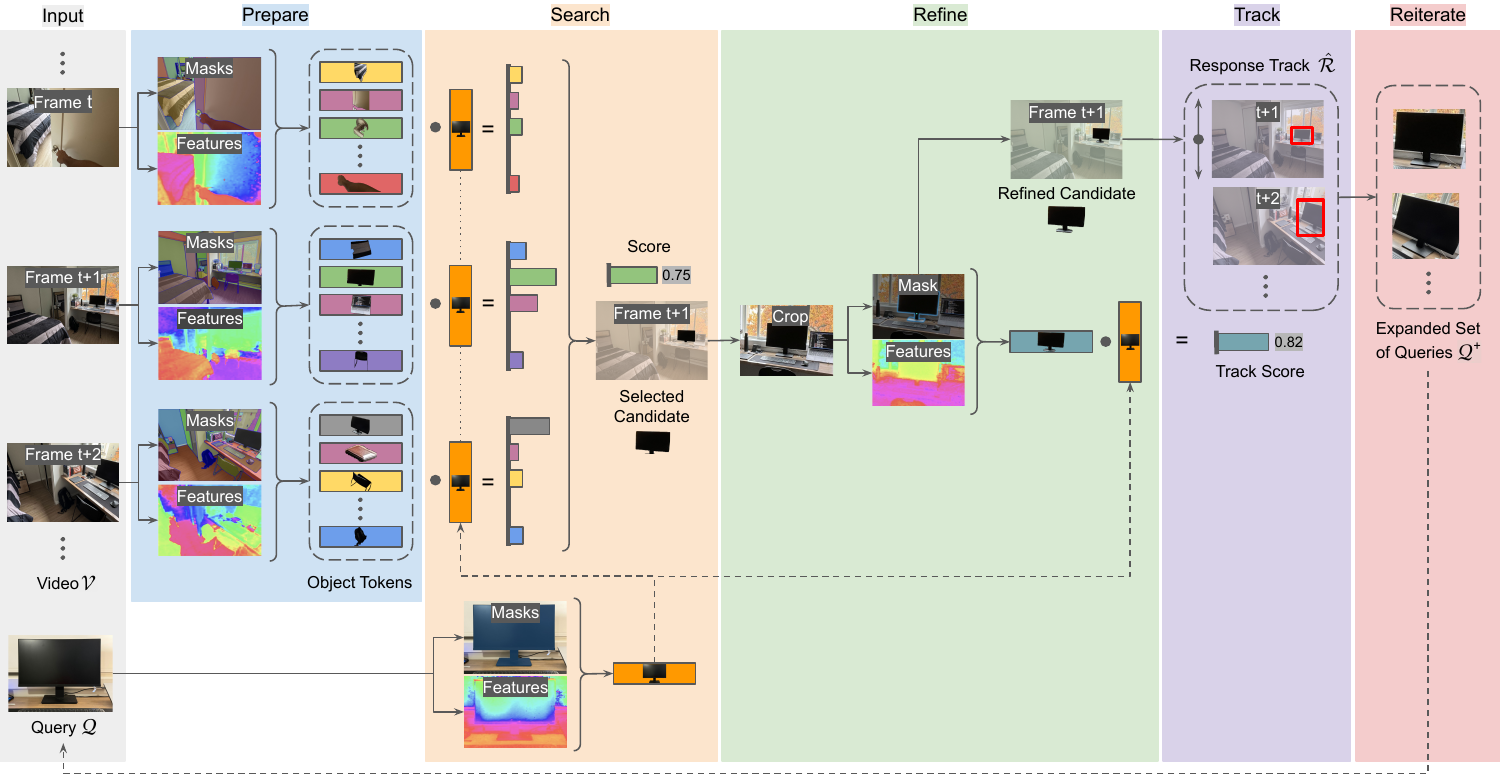}
    \caption{{\bf \vqlb{} framework:} \textcolor{lightblue}{Prepare} object tokens from the given video $\mathcal{V}$ and \textcolor{lightorange}{search} for candidates that match the visual query $\mathcal{Q}$. Then, \textcolor{lightgreen}{refine} the search results for better precision and \textcolor{mauve}{track} the latest refined candidate across video frames to get a response track prediction $\hat{\mathcal{R}}$. Finally, use the response track to create additional visual queries and \textcolor{lightred}{reiterate} the search, refinement, and tracking process.}
    \label{fig:framework}
\end{figure*}

\section{\vqlb{}}
\label{sec:method}
We focus on VQL, the task of localizing the last occurrence of a query object in a video. Formally, given a video $\mathcal{V}$, a visual query $\mathcal{Q}$, and a query time $T$, the objective is to predict a response track $\mathcal{R} = \{b_s, b_{s+1}, \dots, b_e\}$ that localizes and tracks the latest occurrence of the query object prior to time $T$. Here, $\mathcal{Q}$ is specified by a bounding box in a reference frame, $s$ and $e\leq T$ denote the first and the last frames in which the query object is visible, and $b_i$ represents a bounding box around the object in frame $i$.

As shown in Figure~\ref{fig:framework}, \vql{} \textit{prepares} video representations to facilitate a swift \textit{search} for objects that match the given visual queries. It then \textit{refines} the search results for better precision and \textit{tracks} the latest match across  frames. After making a prediction, it can \textit{reiterate} the localization process using the previous prediction for better results.

\boldheader{Prepare (Section~\ref{sec:prepare})}
A given video is first distilled into semantically meaningful object-level representations, or \textit{object tokens}. For this, \vql{} uses a segmentation model to generate object-wise binary masks for each object across all video frames and a feature extractor to produce dense feature maps for each frame. The features within each object mask are then pooled to produce an object token.

\boldheader{Search (Section~\ref{sec:search})}
After preparing object tokens from the video, \vql{} creates a similar region-based representation for the visual query, referred to as the \textit{query token}. It then searches the object tokens for candidates that match the query token.

\boldheader{Refine (Section~\ref{sec:refine})}
The candidates identified in the initial search are then refined to improve spatial precision and remove any spurious matches. For example, Figure~\ref{fig:framework} illustrates how refinement enhances spatial precision by capturing the base of the monitor.

\boldheader{Track (Section~\ref{sec:track})}
The latest refined search is tracked across video frames using an off-the-shelf tracking model to produce a response track localizing the most recent occurrence of the query object in the video.

\boldheader{Reiterate (Section~\ref{sec:reiterate})} 
To better capture the visual variations of the query object, we leverage the object's appearance in the tracked frames to create more visual queries and then re-apply \vql{} to the video segment that follows the previously predicted track.

\subsection{Prepare}
\label{sec:prepare}
To address VQL, it is imperative to encode and represent all regions/objects in each frame of a video. However, creating patch-level features for each frame results in an excessive number of tokens for long videos, making the task computationally expensive. To address this issue, we propose using region-based representations \cite{ShlapentokhRothman2024RegionBasedRR} to encode each frame. As shown in Table~\ref{tab:num-tokens}, the region-based approach significantly reduces the number of tokens needed to encode a video. Moreover, this produces semantically meaningful object tokens, simplifying entity search for a given query.

Building on this insight, \vql{} prepares object tokens from a given video as follows:

\boldheader{1. Segment Objects} 
We first generate a binary mask for each region/object in all frames of the video via a segmentation model. Formally, for a frame $f_t \in \mathbb{R}^{H \times W \times 3}$, we extract a set of binary masks $M_t = \{m_{t1}, m_{t2}, \dots\}$, one for each region. Here, $m_{ti} \in \mathbb{R}^{H \times W}$ is a binary mask with 1 in the area occupied by the object $i$ and 0 otherwise. 
In this work, we use SAM ViT-H~\cite{Kirillov2023SegmentA} as the segmentation model, which processes each frame at a resolution of $1024 \times 1024$.

\boldheader{2. Extract Features} 
We then extract dense features from every frame of the video via a pretrained vision model. Formally, for every frame $f_t \in \mathbb{R}^{H \times W \times 3}$ we compute a high-dimensional feature map $h_t \in \mathbb{R}^{h \times w \times d}$. 
In this work, we experiment with different feature extractors and choose DINO ViT-B/8 \cite{Caron2021EmergingPI} with a frame resolution of $384 \times 512$.

\boldheader{3. Resize and Pool}
Subsequently, we resize the frame features $h_t \in \mathbb{R}^{h \times w \times d}$ to $\overline{h}_t \in \mathbb{R}^{H \times W \times d}$ to fit the height and width of the masks. We then pool the features within each object mask $m_{ti}$ to get an object token $o_{ti} \in \mathbb{R}^{d}$. Following the insights from \cite{ShlapentokhRothman2024RegionBasedRR}, we use bilinear interpolation to resize the features and average pooling to aggregate the features within the mask.

\subsection{Search}
\label{sec:search}
Videos spanning tens of minutes and processed at 5 FPS typically involve more than 150,000 object tokens, with only about 0.01\% corresponding to the target response track. To find candidates similar to the visual query, \vql{} searches over the object tokens. Concretely, given an image and a bounding box localizing the query object, we extract the query token via the process used  for video preparation and proceed with the following steps:

\boldheader{1. Compute Similarity} 
We compute pairwise cosine similarity between all object tokens and query tokens. Initially, only one query token is generated from the given visual query, but multiple query tokens can be created for the same object. When multiple query tokens are present, we use the maximum similarity score across all tokens as the object score, ensuring that an object is selected if it matches any view of the query. Formally, for the $i^\text{th}$ object token in frame $t$ (\ie, $o_{ti}$) and $m$ query tokens $[q_j]_{j=1}^{m}$, the similarity score $s_{ti}$ is computed via  
\begin{equation}
    \vspace{-0.5pt}
    s_{ti} = \max_{j\in\{1,\ldots,m\}} \frac{o_{ti} \cdot q_j}{\|o_{ti}\| \|q_j\|}.
    \label{eq:similarity}
\end{equation}
We also studied test-time training (TTT)~\cite{Sun2019TestTimeTW}, but it performed worse than a simple cosine similarity search due to an insufficient number of query tokens for effective training.

\boldheader{2. Perform Intra-Frame NMS} 
Given that each frame can contain at most one instance of the query object, we apply non-maximum suppression (NMS) to retain only the highest-scoring object per frame, setting the scores of all other objects to zero. This process results in a single candidate object per frame, yielding a total number of candidate objects equal to the number of frames in the video.

\boldheader{3. Perform Inter-Frame NMS} 
Since we perform bidirectional tracking at a later stage, selecting one candidate from a potential track is sufficient. So, we perform NMS across consecutive frames to select the highest-scoring match for a query object, suppressing lower-scoring instances in neighboring frames. More precisely, we iteratively select the object with the maximum score and nullify candidate scores in preceding and subsequent frames until the score drops below 80\% of this peak score. As a result, we end up with sparsely selected high-scoring objects across video frames.

\boldheader{4. Select Candidates} 
Finally, we select up to $k=10$ candidate objects that exceed a $t_{\text{sim}}=0.7$ similarity threshold, yielding an average of 7.5 candidates per query.

\subsection{Refine}
\label{sec:refine}
The objects of interest are often small and situated in cluttered scenes, making it challenging for the segmentation model and feature extractor to produce high-quality object tokens. To address this, we propose refining the selections made after the initial search. We find that this refinement significantly enhances the spatial precision of \vql{} and helps filter out any spurious selections made after the initial search (see Section~\ref{sec:ablation-study} for an ablation). 

We refine the candidates selected from the search in conjunction with the query to generate a more precise set of candidates. This is done as follows:

\boldheader{1. Get Object-Centric Crop} 
We crop the video frames containing the selected candidates and visual queries such that these objects occupy a larger area at the center of the cropped view. These crops are then resized to the original frame dimensions. To prevent pixelation in the case of extremely small objects, we ensure that no frame is cropped with a zoom factor exceeding 2.5 times.

\boldheader{2. Generate Refined Token} 
We process the object-centric crops of candidates and queries using the segmentation model and feature extractor to generate refined object and query tokens. This process yields improved tokens because the segmentation model and feature extractor can produce more accurate masks and detailed features for the objects from the object-centric crop.

\boldheader{3. Recompute Similarity and Filter}
We recompute the cosine similarity scores for the selected candidates using the refined object tokens and query tokens, and filter out  candidates that have a score below  threshold $t_{\text{sim}}=0.7$.

\subsection{Track}
\label{sec:track}
\vql{} uses an off-the-shelf tracker to bidirectionally track the refined candidate that shows up last in the video, yielding the response track $\mathcal{\hat{R}} = \{b_s, b_{s+1}, \dots, b_e\}$. Further, it assigns the similarity score of the candidate as the track score. In this work, we use SAM 2~\cite{Ravi2024SAM2S} for tracking.

\subsection{Reiterate}
\label{sec:reiterate}
The query object often appears multiple times in a given video, and \vql{} is highly successful at localizing at least one of these appearances from a single visual query (see Section~\ref{sec:vq2d-evaluation}). However, our goal is to identify the {\em latest} appearance, which may be heavily occluded or seen from a significantly different viewpoint compared to the visual query. To achieve this, we generate additional visual queries using the response track predicted with the original query and reiterate the process of search, refinement, and tracking. This expanded pool of queries offers diverse views of the object, enhancing the likelihood of detection even when it is obscured or appears in a radically altered form. In practice, we perform this query expansion and reiteration step only once. This is carried out as follows:

\boldheader{1. Generate Query Tokens} 
Given the response track $\mathcal{\hat{R}} = \{b_s, b_{s+1}, \dots, b_e\}$, we apply the segmentation model and feature extractor to produce region tokens for the objects within the bounding boxes across the frames comprising the response track.

\boldheader{2. Filter Queries} 
We filter out low-quality query tokens generated in the previous step, specifically targeting three types: (1) query tokens with very low cosine similarity (less than 0.5) to the original query token, (2) queries associated with extremely small bounding boxes (occupying less than 0.07\% of the frame area), and (3) queries derived from blurry frames (indicated by a Laplacian operator variance below 100). 

\boldheader{3. Search, Refine, and Track}
After expanding the query pool, we search the video segment following the last frame of the previously predicted response track. The search results are then refined, and the latest refined candidate is tracked across frames to generate a new prediction. If the score of the new prediction exceeds a threshold relative to the previous track score, we update the previous prediction.

%% file: sections/4-experiments.tex
\section{Experiments}
\label{sec:experiments}
In this section, we evaluate \vql{} (Section~\ref{sec:vq2d-evaluation}), discuss key design decisions (Section~\ref{sec:design-decisions}), and present ablation studies (Section~\ref{sec:ablation-study}).

\input{tables/vq2d-evaluation}

\subsection{Evaluation}
\label{sec:vq2d-evaluation}
\boldheader{Dataset}
We evaluate \vql{} on the Ego4D VQ2D dataset~\cite{Grauman2021Ego4DAT}, a large collection of egocentric videos annotated for VQL within episodic memory. On average, the videos in this dataset are 140 seconds long, and the target response tracks last for roughly 3 seconds. The dataset comprises 13,600 training, 4,500 validation, and 4,400 test queries, annotated across 262, 87, and 84 hours of video, respectively. We use the validation set for our development and ablations. To our knowledge, VQ2D is the only publicly available dataset for VQL. 

\boldheader{Metrics}
Following the official metrics outlined by the benchmark, we report spatio-temporal average precision ($\text{stAP}_{25}$), temporal average precision ($\text{tAP}_{25}$), success, and recovery. $\text{stAP}_{25}$ and $\text{tAP}_{25}$ evaluate the accuracy of the predicted response tracks' spatio-temporal and temporal extents using an Intersection over Union (IoU) threshold of 0.25. Success measures whether the IoU between predictions and ground truth exceeds 0.05, and recovery measures the proportion of predicted frames where the bounding box achieves an IoU of at least 0.5 with the ground truth.

\boldheader{Results}
Table~\ref{tab:vq2d-evaluation} shows results on the validation and test sets. Despite no task-specific training, \vql{} outperforms the next-best baseline by 49\% $\text{stAP}_{25}$, 33\% $\text{tAP}_{25}$, 8\% Success, and 12\% Recovery on the test set. Note, these are relative improvements. Figure~\ref{fig:teaser} shows some qualitative examples of \vql{} in action.

\begin{figure}[t]
    \captionsetup{skip=5pt}
    \centering
    \begin{subfigure}[t]{0.47\textwidth}
        \centering
        \includegraphics[width=\linewidth]{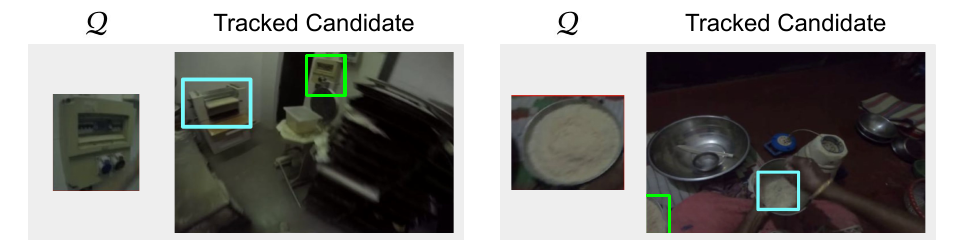}
        \caption{An object visually similar to the query (blue boxes) is mistakenly tracked instead of the correct query object (green boxes).}
        \label{fig:failure-resemble}
    \end{subfigure}
    \vspace{0.1cm}
    \begin{subfigure}[t]{0.47\textwidth}
        \centering
        \includegraphics[width=\linewidth]{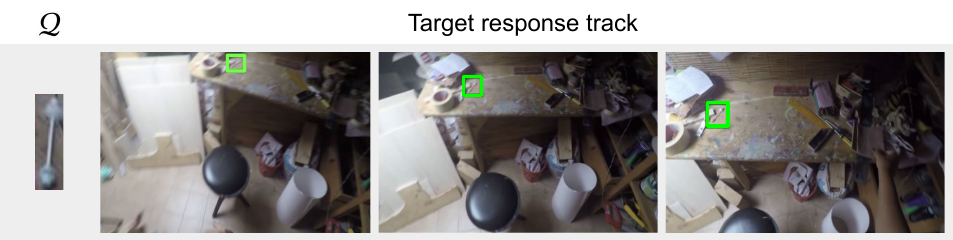}
        \caption{The visual query shows the target object from a distinctly different viewpoint compared to its single appearance in the video (green boxes).}
        \label{fig:failure-view}
    \end{subfigure}
    \vspace{0.1cm}
    \begin{subfigure}[t]{0.47\textwidth}
        \centering
        \includegraphics[width=\linewidth]{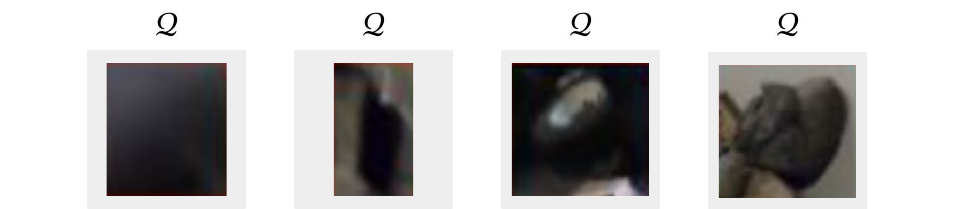}
        \caption{These visual queries lack discriminative features due to motion blur, low contrast with background, or poor visibility.}
        \label{fig:failure-query}
    \end{subfigure}
    \caption{{\bf Examples where a wrong object is localized.} Among 100 randomly sampled instances from the VQ2D validation set, \vql{} localized an instance of the query object in 93 cases, and failed to localize the correct object only under extreme conditions.}
    \label{fig:error-analysis}
\end{figure}

Additionally, a manual analysis of 100 randomly sampled examples from the VQ2D validation set shows that \vql{} successfully localizes the last occurrence of the object in 61 cases, localizes an earlier instance in 32 cases, and identifies the wrong object in 7 cases. Figure~\ref{fig:error-analysis} categorizes these failure modes for the cases where \vql{} localizes the wrong object, highlighting that the framework typically fails to localize the correct object only in extreme situations, such as when the visual query is ambiguous or there exists another object that closely resembles the query.

\boldheader{Timing Analysis} 
On average, preparing a 1000-frame video (\ie, extracting object tokens) takes 1422.5 seconds with our current setup. Subsequent steps are faster: search takes 0.8 seconds, refinement 12.6 seconds, tracking 26.3 seconds, and generating additional queries from a response track is instantaneous. Note that our current implementation is not optimized for speed. For applications requiring faster localization, several simple optimizations can be made with minimal impact on the metrics, such as using HNSW~\cite{Malkov2016EfficientAR} for search, increasing the batch size for feature extraction, using faster SAM variants~\cite{Pytorch2023Segment} for region generation, and employing faster trackers. We leave these enhancements for future work.

\subsection{Design Decisions}
\label{sec:design-decisions}
\boldheader{Region vs.\ Patch Representations}
Most contemporary approaches to object localization and tracking encode video frames using patch-based representations, where each frame is divided into non-overlapping patches and encoded with a vision transformer. Region-based representations derived from pretrained vision models \cite{ShlapentokhRothman2024RegionBasedRR} have recently been proposed as an efficient alternative. These representations offer both semantic richness and compact encoding—qualities that align  with the requirements of VQL.

Notably, region token count depends on the scene content (detected objects or regions) rather than spatial parameters (image dimensions and patch size). As a result, in most practical cases, a region-based approach yields a much lower token count than the patch-based approach. As shown in Table~\ref{tab:num-tokens}, a region-based encoding of a 200-second VQ2D video produces $24 \times$ fewer tokens compared to a patch-based encoding at an image resolution of $384 \times 512$ with a patch size of 8. Moreover, this difference drastically increases if we increase the resolution; \eg if the image resolution is increased to $1024 \times 1024$, a region-based encoding produces $126 \times$ fewer tokens than a patch-based encoding. Thus, the decoupling of token count from spatial parameters allows the region-based approach to leverage the benefits of smaller patches and high-resolution images without incurring the computational overhead typically associated with increased token counts.

\begin{figure}[t]
    \captionsetup{skip=5pt}
    \centering
    \includegraphics[width=0.99\linewidth]{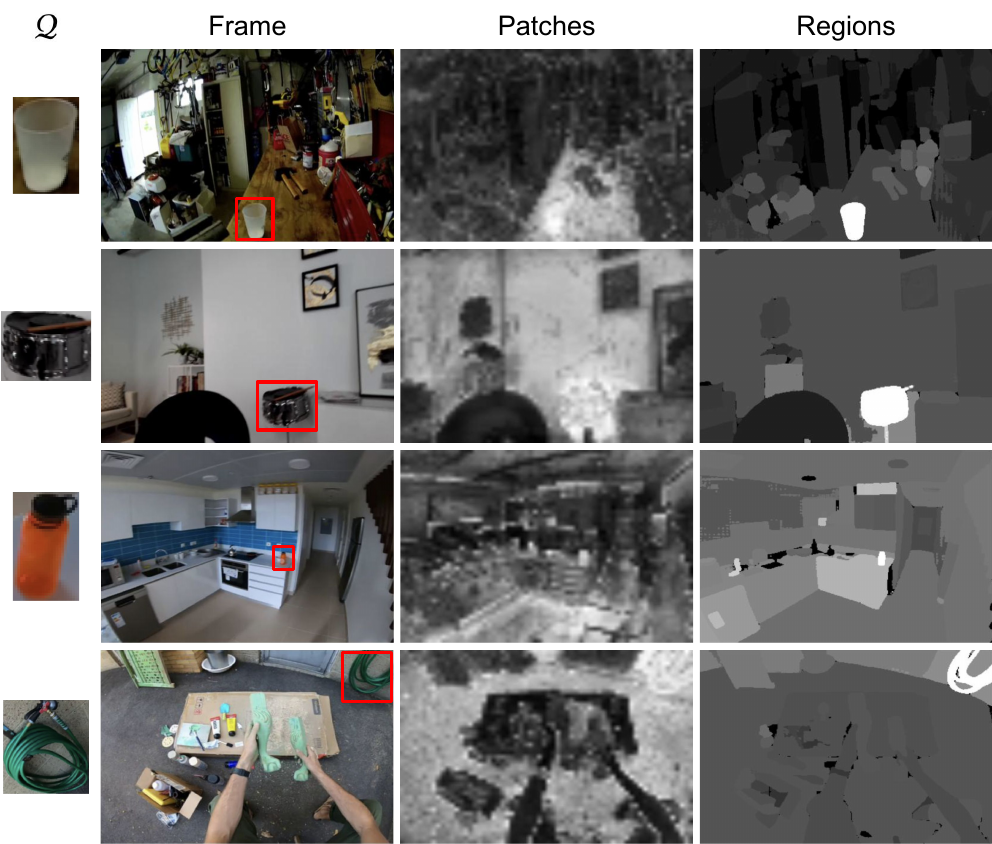}
    \caption{{\bf Comparing cosine similarity heatmaps for patch-based and region-based representations.} Region-based representations yield distinct, high-similarity matches for query objects (brighter regions), while patch-based representations produce more diffuse similarity patterns.}
    \label{fig:similarity-heatmap}
\end{figure}

Beyond their compactness, region-based representations excel at capturing semantic information, enabling spatio-temporal object localization through simple token similarity matching. Figure~\ref{fig:similarity-heatmap} illustrates this through cosine similarity heatmaps, where region-based representations show precise areas of high similarity for the query objects, while patch-based representations result in less focused similarity distributions. Our comparative analysis reveals that while patch encodings achieve better frame-level retrieval (65\% vs.\ 60\%), they perform significantly worse at spatial localization (42\% vs.\ 58\%) compared to region-based representations. For specific applications, one could combine patch-based frame retrieval with region-based object localization. For episodic memory tasks, however, we believe that a streamlined approach using only region-based representations is advantageous as it avoids the computational overhead of storing and searching through substantially more patch tokens and eliminates the need to generate regions during retrieval.

\input{tables/num-tokens}

\boldheader{Feature Extraction}
To search and retrieve objects that match a given query, VQL requires capturing fine-grained details from each video frame, and the DINO models are particularly well-suited for this task \cite{ShlapentokhRothman2024RegionBasedRR, Jiang2023FromCT}. We compare DINO~\cite{Caron2021EmergingPI} and DINOv2~\cite{Oquab2023DINOv2LR} using various ViT backbones, with results shown in Table~\ref{tab:extractor-comparison}. Our findings indicate that DINO ViT-B/8 outperforms ViT-B/16, likely due to its smaller patch size enabling finer feature extraction. Additionally, DINOv2 ViT-L/14 shows stronger frame-level retrieval, and DINO ViT-B/8 is better at spatial localization of objects. We choose DINO ViT-B/8 as our feature extractor due to its higher spatio-temporal localization success rate, smaller model size for faster inference, and reduced feature dimensionality requiring less memory.

\input{tables/extractor-comparision}

\begin{figure*}[t]
    \captionsetup{skip=5pt}
    \centering
    \begin{subfigure}[t]{0.46\textwidth}
        \centering
        \includegraphics[width=\linewidth]{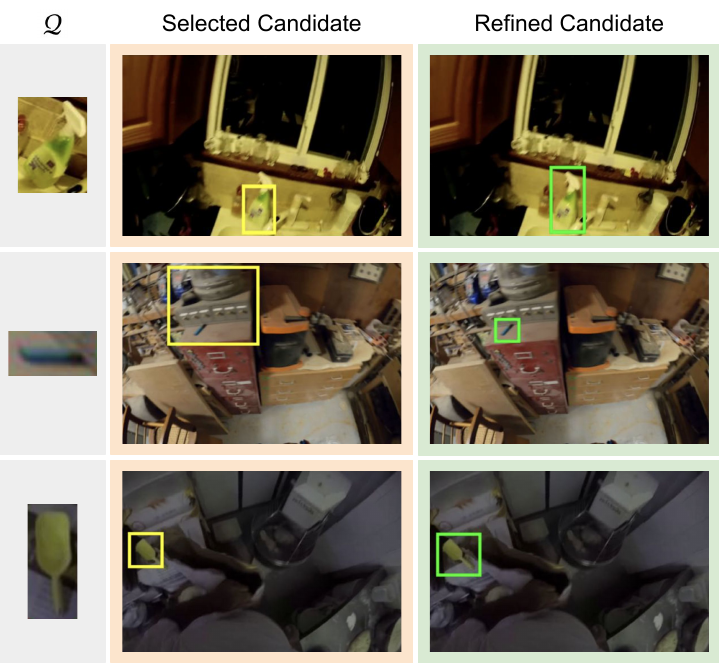}
        \caption{Cropping around objects during refinement helps capture small objects or object parts that full-frame processing might miss.}
        \label{fig:refinement-spatial}
    \end{subfigure}
    \hspace{0.02\textwidth}
    \begin{subfigure}[t]{0.46\textwidth}
        \centering
        \includegraphics[width=\linewidth]{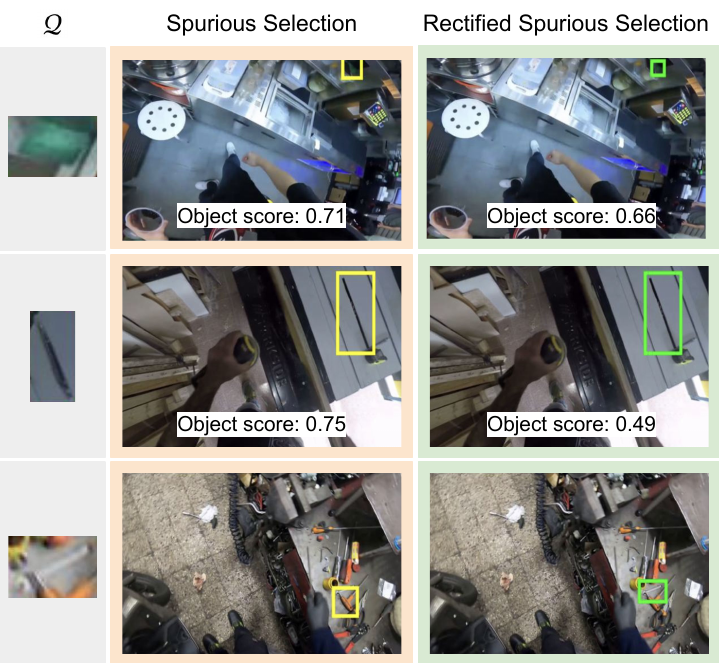}
        \caption{Refinement fixes incorrect selections due to feature bleeding (row 1) and removes false selections by lowering their scores (rows 2 and 3).}
        \label{fig:refinement-spurious}
    \end{subfigure}
    \vspace{-0.05cm}
    \caption{{\bf Impact of refining search results.} Initial candidates are shown in yellow boxes, and refined candidates are shown in green.}
    \label{fig:refinement-effect}
\end{figure*}

\begin{figure*}[t]
    \captionsetup{skip=3pt}
    \centering
    \includegraphics[width=0.93\linewidth]{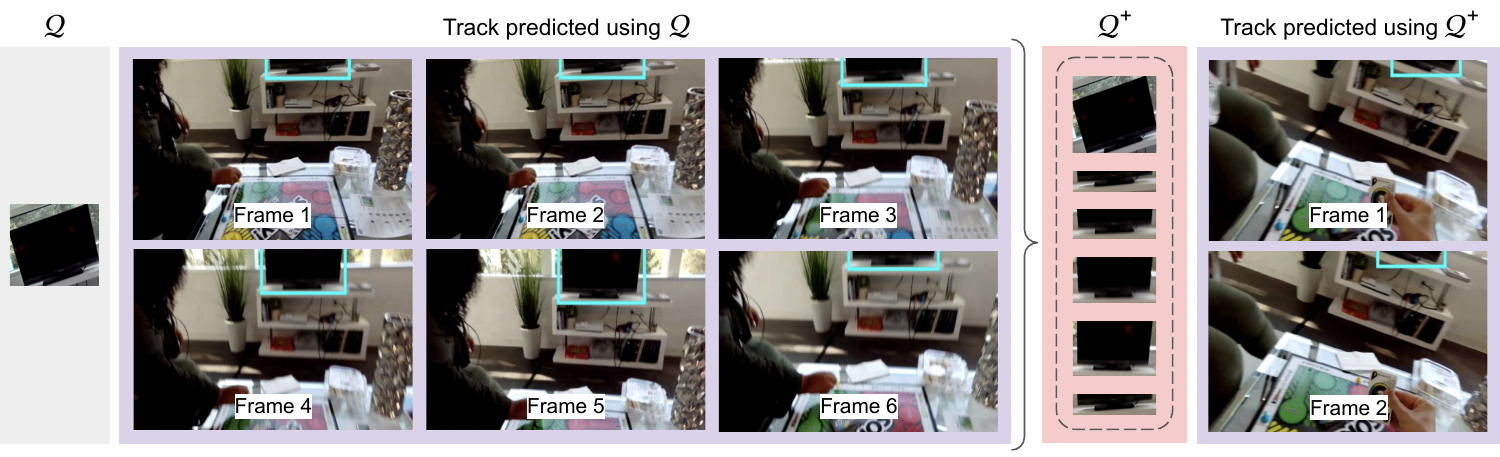}
    \caption{{\bf Impact of query expansion and reiteration.} Using only the initial query $\mathcal{Q}$, \vql{} misses the object's final appearance due to its partial view and low feature similarity. However, the expanded query set $\mathcal{Q}^{+}$ captures multiple object perspectives, enabling successful detection of the final occurrence.}
    \label{fig:reiteration-effect}
\end{figure*}

\input{tables/ablation-study}

\subsection{Ablation Study}
\label{sec:ablation-study}
\boldheader{Refining Search Results}
Results in Table~\ref{tab:ablation-study} show that removing search refinement (\vql{}-NoRefine) from the framework significantly degrades \vql{}'s performance. As illustrated in Figure~\ref{fig:refinement-spatial}, refinement enhances spatial precision for localizing small objects and fine parts by generating regions from an object-centric crop that captures more of the object of interest. Additionally, refinement removes incorrect selections from the initial search by rescoring candidates using region tokens generated from an object-centric crop. Furthermore, for small or thin objects, feature bleeding can result in the selection of an object adjacent to the target object. Object-centered cropping can also help avoid this by providing a clearer view of the target. This is demonstrated in Figure~\ref{fig:refinement-spurious}.

\boldheader{Query Expansion and Reiteration}
Results in Table~\ref{tab:ablation-study} show that query expansion and reiteration significantly improve \vql{}'s performance compared to the variant without this step (\vql{}-NoReiter). Figure~\ref{fig:reiteration-effect} shows an example where query expansion using an initial prediction helps capture a wider range of views for the visual query. This, in turn, helps in successfully localizing the final occurrence of the object that was missed when only the initial visual query was used. Additional examples where reiteration after query expansion leads to successful localization are illustrated in rows 2, 5, and 6 of Figure~\ref{fig:teaser}.

%% file: tables/vq2d-evaluation.tex
\begin{table}[t]
    \captionsetup{skip=5pt}
    \centering
    \footnotesize
    \begin{tabular}{c c c c c}
        \toprule
        \textbf{Method} & $\mathbf{\textbf{stAP}_{25}}$ & $\mathbf{\textbf{tAP}_{25}}$ & \textbf{Success} & \textbf{Recovery} \\
        \midrule
        \multicolumn{5}{c}{\textit{Validation Set}} \\
        \midrule
        SiamRCNN~\cite{Grauman2021Ego4DAT} & 0.15 & 0.22 & 43.2 & 32.9 \\
        NFM~\cite{Xu2022NegativeFM} & 0.19 & 0.26 & 47.9 & 37.9 \\
        CocoFormer~\cite{Xu2022WhereIM} & 0.19 & 0.26 & 47.7 & 37.7 \\
        VQLoC~\cite{Xu2022WhereIM} & 0.22 & 0.31 & 55.9 & 47.1 \\
        \vqlb{} & \textbf{0.33} & \textbf{0.41} & \textbf{58.0} & \textbf{50.5} \\
        \midrule
        \multicolumn{5}{c}{\textit{Test Set}} \\
        \midrule
        SiamRCNN~\cite{Grauman2021Ego4DAT} & 0.13 & 0.21 & 41.6 & 34.0 \\
        CocoFormer~\cite{Xu2022WhereIM} & 0.18 & 0.26 & 48.1 & 43.2 \\
        VQLoC~\cite{Xu2022WhereIM} & 0.24 & 0.32 & 55.9 & 45.1 \\
        \vqlb{} & \textbf{0.35} & \textbf{0.43} & \textbf{60.1} & \textbf{50.6} \\
        \bottomrule
    \end{tabular}
    \caption{{\bf Results on Ego4D VQ2D benchmark.} The validation results are taken from \cite{Xu2022WhereIM}, and the test results are obtained from the \href{https://eval.ai/web/challenges/challenge-page/1843/leaderboard/4326}{challenge leaderboard}.}
    \label{tab:vq2d-evaluation}
\end{table}

%% file: tables/num-tokens.tex
\begin{table}[t]
    \captionsetup{skip=5pt}
    \centering
    \footnotesize
    \begin{tabular}{c c c}
        \toprule
        \multirow{2}{*}{\textbf{Encoding Method}} & \multicolumn{2}{c}{\textbf{Number of tokens}} \\
        \cmidrule(lr){2-3}
        & \textbf{384 $\times$ 512 Frames} & \textbf{1024 $\times$ 1024 Frames} \\
        \midrule
        Patches & 3,072,000 & 16,384,000 \\
        \textbf{Regions} & \textbf{130,507} & \textbf{130,507} \\
        \bottomrule
    \end{tabular}
    \caption{{\bf Token count for patch-based and region-based encodings.} Region-based representations typically use fewer tokens than the patch-based method, and their token count does not scale with resolution. Token counts are reported for a 1000-frame video, using a patch size of 8 and SAM ViT-H for region-based encoding.}
    \label{tab:num-tokens}
\end{table}

%% file: tables/extractor-comparision.tex
\begin{table}[t]
    \captionsetup{skip=5pt}
    \centering
    \footnotesize
    \begin{tabular}{c c c c c c}
        \toprule
        \textbf{Extractor} & $\mathbf{\textbf{stAP}_{25}}$ & $\mathbf{\textbf{tAP}_{25}}$ & \textbf{Success} & \textbf{Recovery} \\
        \midrule
        DINO ViT-B/16 & 0.254 & 0.301 & 50.0 & 47.7 \\
        DINOv2 ViT-L/14 & \textbf{0.331} & \textbf{0.452} & 55.8 & 45.3 \\
        \textbf{DINO ViT-B/8} & \textbf{0.333} & 0.409 & \textbf{58.0} & \textbf{50.5} \\
        \bottomrule
    \end{tabular}
    \caption{{\bf Performance comparison of feature extractors.} DINOv2 ViT-L/14 shows superior frame-level retrieval while DINO ViT-B/8 is better at spatial localization.}
    \label{tab:extractor-comparison}
\end{table}

%% file: tables/ablation-study.tex
\begin{table}[t]
    \captionsetup{skip=5pt}
    \centering
    \footnotesize
    \begin{tabular}{c c c c c}
        \toprule
        \textbf{Method} & $\mathbf{\textbf{stAP}_{25}}$ & $\mathbf{\textbf{tAP}_{25}}$ & \textbf{Success} & \textbf{Recovery} \\
        \midrule
        \vql{}-NoRefine &  0.246 & 0.364 & 49.1 & 39.8 \\
        \vql{}-NoReiter & 0.247 & 0.293 & 50.9 & 48.2 \\
        \vqlb{} & \textbf{0.333} & \textbf{0.409} & \textbf{58.0} & \textbf{50.5} \\
        \bottomrule
    \end{tabular}
    \caption{{\bf Impact of refining search results and reiterating after query expansion.} Ablating either refinement (\vql{}-NoRefine) or reiteration (\vql{}-NoReiter) significantly reduces the performance of \vql{}.}
    \label{tab:ablation-study}
\end{table}

%% file: sections/5-conclusion.tex
\section{Conclusion}
\label{sec:conclusion}
We present \vql{}, a framework utilizing region-based representations from pretrained vision models to address VQL. Despite a simple design and no task-specific training, \vql{} can localize target objects in long videos, even under challenging conditions such as clutter, occlusion, blur, and viewpoint changes. On the VQ2D benchmark, it significantly outperforms existing methods that are specifically trained on this dataset. 

%% file: sections/6-supplementary.tex
\appendix
\setcounter{page}{1}
\maketitlesupplementary

\noindent{}This supplementary material is structured as follows. In \cref{sec:hyperparam-analysis} we analyze the sensitivity of \vql{} to its hyperparameters. In \cref{sec:sam2-vq2d} we study the performance of SAM 2 on the VQL task. 

\section{Hyperparameter Sensitivity Analysis}
\label{sec:hyperparam-analysis}
\begin{figure*}[t]
    \centering
    \begin{subfigure}[b]{0.48\textwidth}
        \centering
        \includegraphics[width=\textwidth]{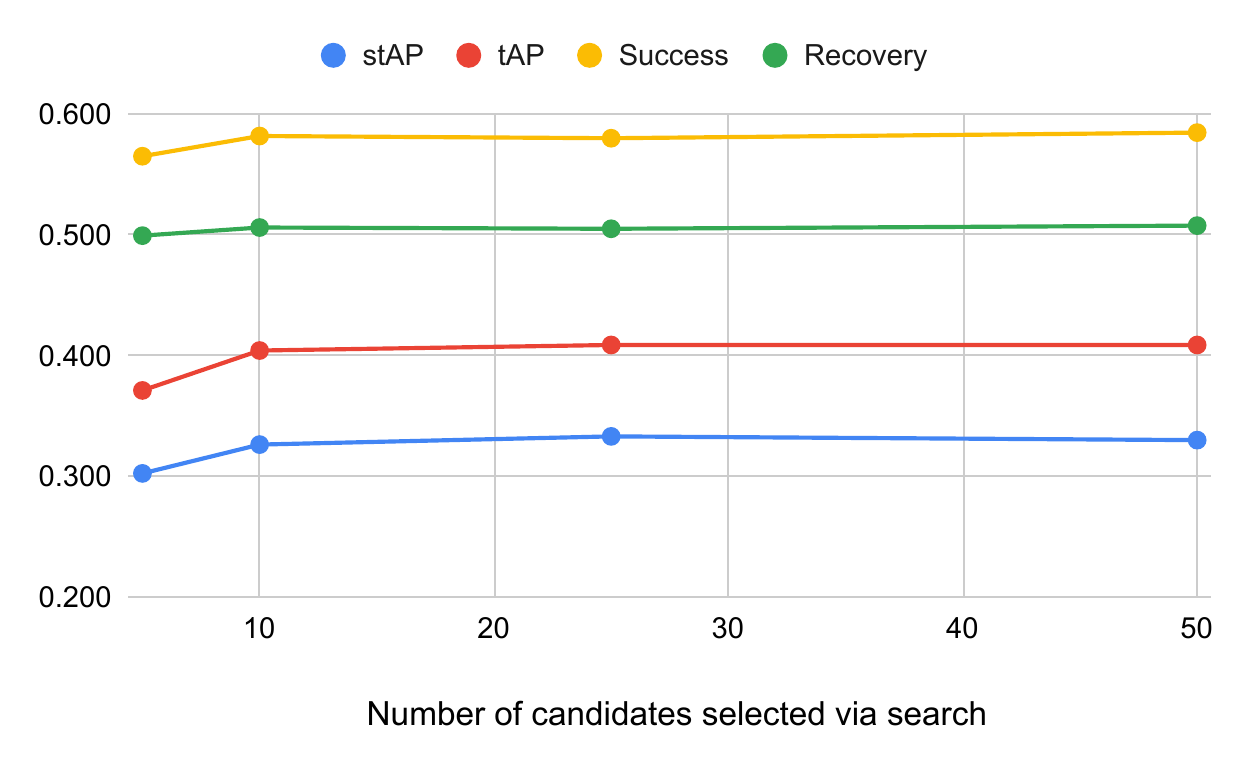}
        \label{fig:hyperparam-k}
    \end{subfigure}
    \hfill
    \begin{subfigure}[b]{0.48\textwidth}
        \centering
        \includegraphics[width=\textwidth]{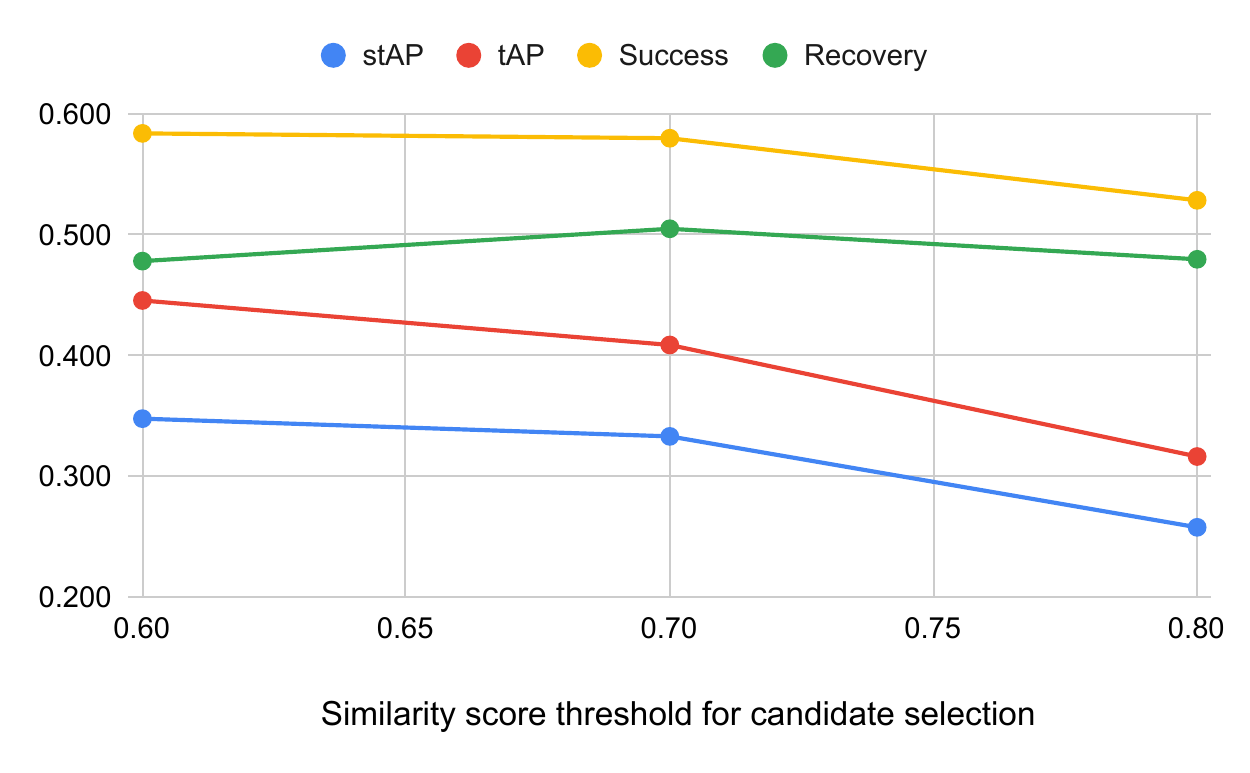}
        \label{fig:hyperparam-tsim}
    \end{subfigure}
    \begin{subfigure}[b]{0.48\textwidth}
        \centering
        \includegraphics[width=\textwidth]{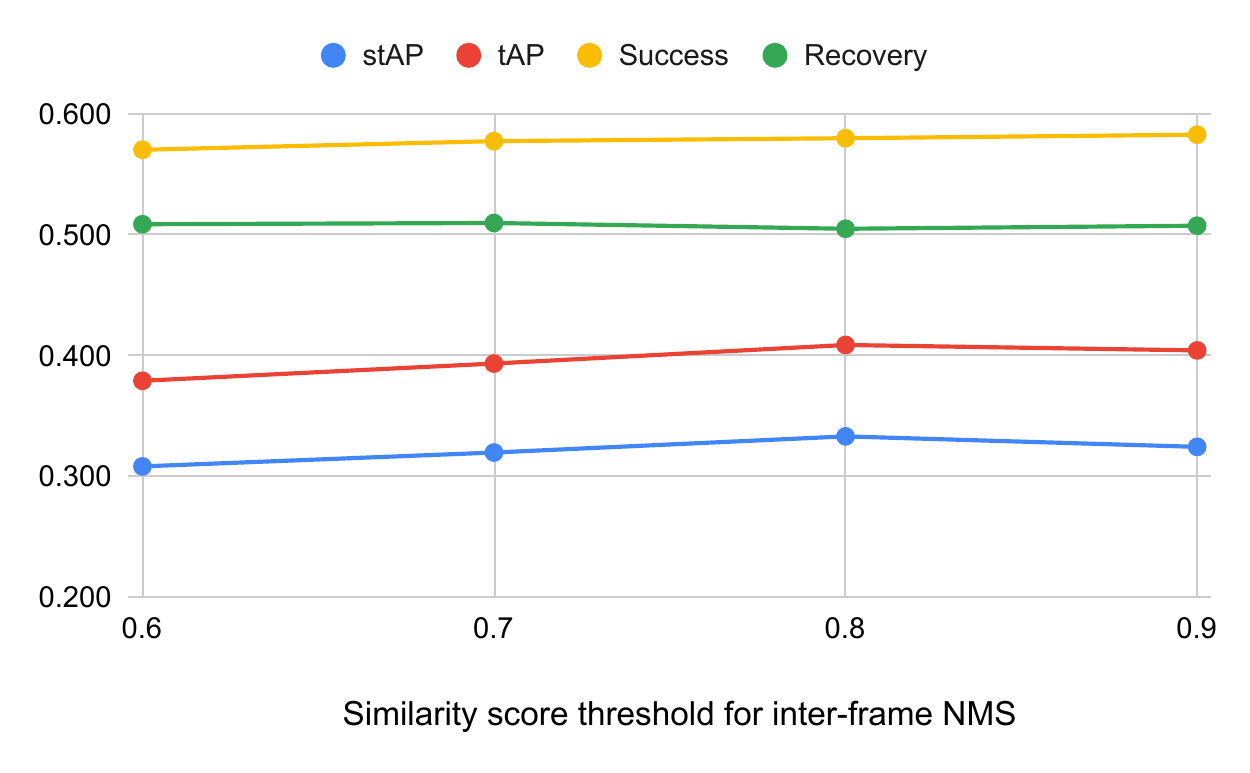}
        \label{fig:hyperparam-nms}
    \end{subfigure}
    \hfill
    \begin{subfigure}[b]{0.48\textwidth}
        \centering
        \includegraphics[width=\textwidth]{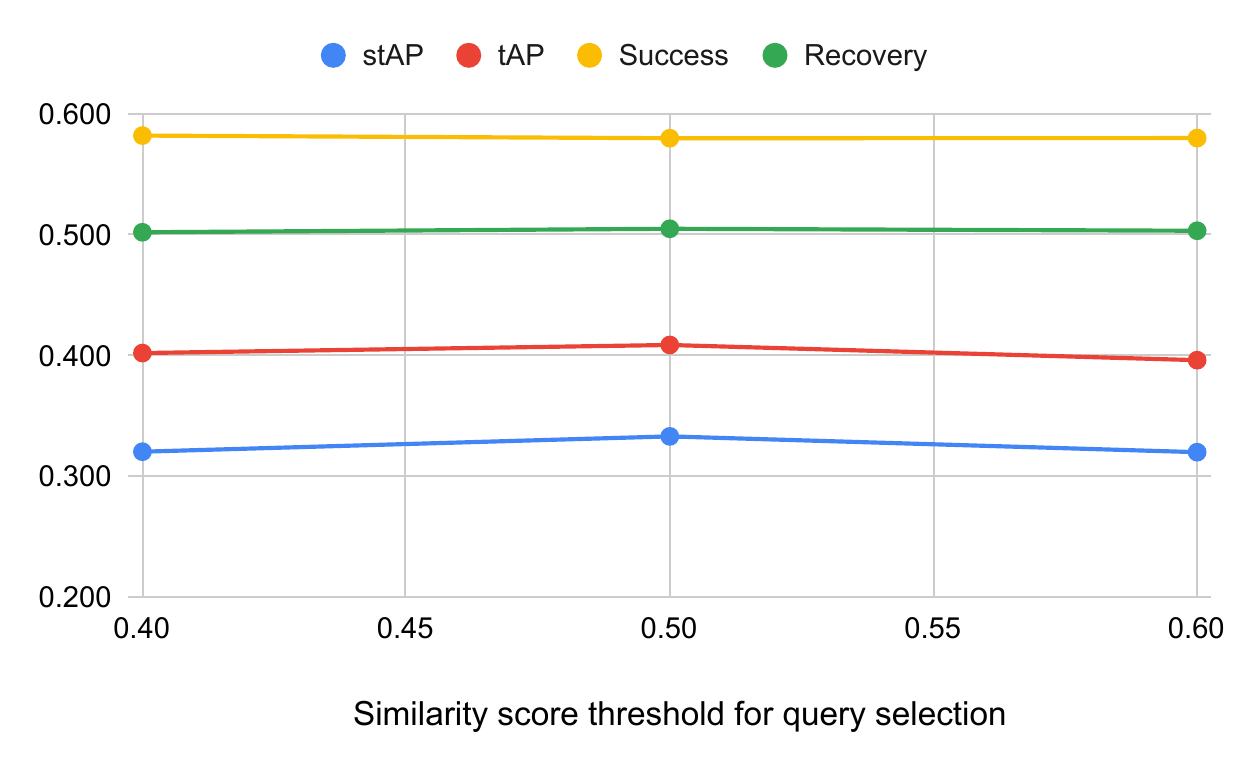}
        \label{fig:hyperparam-query}
    \end{subfigure}
    \caption{{\bf Hyperparameter sensitivity analysis of \vql{}.} Empirical evaluation demonstrates \vql{}'s robustness across different hyperparameter configurations.}
    \label{fig:hyperparam-analysis}
\end{figure*}

We analyze \vql{}'s sensitivity to four key hyperparameters: (1) the maximum number of initially retrieved candidates $k$, (2) the candidate selection threshold $t_\text{sim}$, (3) the inter-frame NMS threshold $t_\text{nms}$, and (4) the query selection threshold $t_\text{q}$. Tables~\ref{tab:hyperparam-k}-\ref{tab:hyperparam-query} and Figure~\ref{fig:hyperparam-analysis} present model's performance across different hyperparameter configurations.

For the initial retrieval count $k$, we observe stable performance across values from 10 to 50, with only a slight degradation at $k=5$. The candidate selection threshold $t_\text{sim}$ leads to a noticeable decline in performance when set above 0.7. The inter-frame NMS threshold $t_\text{nms}$ demonstrates consistent performance across the range 0.7-0.9, suggesting robustness to this parameter. Similarly, the query selection threshold $t_{q}$ shows minimal variation in performance between 0.4 and 0.6.

Overall, these results indicate that our model maintains stable performance across a wide range of hyperparameter values, with selected values of $k=10$, $t_\text{sim}=0.7$, $t_\text{nms}=0.8$, and $t_\text{q}=0.5$ providing a robust operating point.

\input{tables/hyperparam-analysis}
\input{tables/sam2-vq2d}

\section{Evaluating SAM 2 on VQ2D}
\label{sec:sam2-vq2d}
\citet{Jiang2023SingleStageVQ} demonstrated significant limitations in VQL capabilities among contemporary tracking systems. Specifically, they showed that STARK~\cite{Yan2021LearningST}, a state-of-the-art visual tracker at the time, achieves only a 0.04 $\text{stAP}_{25}$ score on the VQ2D validation set. Since then, tracking systems have advanced considerably. To evaluate the capabilities of current tracking systems, we test SAM 2~\cite{Ravi2024SAM2S} on the VQL task. 

To adapt SAM 2 for VQ2D, we prepend the query frame to the target video and use the query bounding box from the annotations as the prompt for mask generation. SAM 2 then propagates the generated mask across all subsequent frames, tracking multiple occurrences of the query object. We select the last contiguous track as the response track prediction.

We evaluate SAM 2 on 100 randomly sampled examples previously used for the manual analysis of \vql{} reported in Section~\ref{sec:vq2d-evaluation}, and the results are shown in Tables~\ref{tab:sam2-vq2d} and \ref{tab:sam2-analysis}. While SAM 2 shows competitive performance on VQ2D (Table~\ref{tab:sam2-vq2d}), it underperforms compared to \vql{}. Our qualitative analysis (Table~\ref{tab:sam2-analysis}) reveals that SAM 2 has a higher tendency to localize incorrect objects or produce no tracks compared to \vql{}. On an NVIDIA A40, with our implementation, SAM 2 takes an average of 110.7 seconds to locate a query object in a 1000-frame video. In comparison, \vql{} incurs a one-time cost of 1422.5 seconds to prepare a 1000-frame video, followed by 73.6 seconds to process each query. However, the processing time of \vql{} can be significantly reduced by using batch processing and faster SAM variants.

%% file: tables/hyperparam-analysis.tex
\begin{table}[t]
    \captionsetup{skip=5pt}
    \centering
    \footnotesize
    \begin{tabular}{c c c c c}
        \toprule
        $\mathbf{k}$ & $\mathbf{\textbf{stAP}_{25}}$ & $\mathbf{\textbf{tAP}_{25}}$ & \textbf{Success} & \textbf{Recovery} \\
        \midrule
        5 &  0.302 & 0.371 & 56.5 & 49.9 \\
        \rowcolor{cyan!15}
        10 & 0.333 & 0.409 & 58.0 & 50.5 \\
        25 & 0.329 & 0.404 & 58.2 & 50.6 \\
        50 & 0.330 & 0.409 & 58.5 & 50.8 \\
        \bottomrule
    \end{tabular}
    \caption{{\bf Effect of initially selected candidates on model performance.} Our final evaluations use $k=10$.}
    \label{tab:hyperparam-k}
\end{table}

\begin{table}[t]
    \captionsetup{skip=5pt}
    \centering
    \footnotesize
    \begin{tabular}{c c c c c}
        \toprule
        $\mathbf{t_{sim}}$ & $\mathbf{\textbf{stAP}_{25}}$ & $\mathbf{\textbf{tAP}_{25}}$ & \textbf{Success} & \textbf{Recovery} \\
        \midrule
        0.6 &  0.348 & 0.446 & 58.4 & 47.8 \\
        \rowcolor{cyan!15}
        0.7 & 0.333 & 0.409 & 58.0 & 50.5 \\
        0.8 & 0.258 & 0.316 & 52.9 & 48.0 \\
        \bottomrule
    \end{tabular}
    \caption{{\bf Effect of candidate selection threshold on model performance.} Our final evaluations use $t_\text{sim}=0.7$.}
    \label{tab:hyperparam-tsim}
\end{table}

\begin{table}[t]
    \captionsetup{skip=5pt}
    \centering
    \footnotesize
    \begin{tabular}{c c c c c}
        \toprule
        $\mathbf{t_{nms}}$ & $\mathbf{\textbf{stAP}_{25}}$ & $\mathbf{\textbf{tAP}_{25}}$ & \textbf{Success} & \textbf{Recovery} \\
        \midrule
        0.6 &  0.308 & 0.379 & 57.1 & 50.9 \\
        0.7 &  0.320 & 0.393 & 57.8 & 51.0 \\
        \rowcolor{cyan!15}
        0.8 & 0.333 & 0.409 & 58.0 & 50.5 \\
        0.9 & 0.324 & 0.404 & 58.3 & 50.8 \\
        \bottomrule
    \end{tabular}
    \caption{{\bf Effect of inter-frame NMS threshold on model performance.} Our final evaluations use $t_\text{nms}=0.8$.}
    \label{tab:hyperparam=nms}
\end{table}

\begin{table}[t]
    \captionsetup{skip=5pt}
    \centering
    \footnotesize
    \begin{tabular}{c c c c c}
        \toprule
        $\mathbf{t_{q}}$ & $\mathbf{\textbf{stAP}_{25}}$ & $\mathbf{\textbf{tAP}_{25}}$ & \textbf{Success} & \textbf{Recovery} \\
        \midrule
        0.4 &  0.320 & 0.402 & 58.2 & 50.2 \\
        \rowcolor{cyan!15}
        0.5 & 0.333 & 0.409 & 58.0 & 50.5 \\
        0.6 & 0.320 & 0.396 & 58.0 & 50.4 \\
        \bottomrule
    \end{tabular}
    \caption{{\bf Effect of query selection threshold on model performance.} Our final evaluations use $t_\text{q}=0.5$.}
    \label{tab:hyperparam-query}
\end{table}

%% file: tables/sam2-vq2d.tex
\begin{table}[t]
    \captionsetup{skip=5pt}
    \centering
    \footnotesize
    \begin{tabular}{c c c c c}
        \toprule
        \textbf{Method} & $\mathbf{\textbf{stAP}_{25}}$ & $\mathbf{\textbf{tAP}_{25}}$ & \textbf{Success} & \textbf{Recovery} \\
        \midrule
        SAM 2~\cite{Ravi2024SAM2S} & 0.290 & 0.329 & 55.0 & 42.7 \\
        \vqlb{} & \textbf{0.378} & \textbf{0.458} & \textbf{63.0} & \textbf{49.1} \\
        \bottomrule
    \end{tabular}
    \caption{{\bf Evaluating SAM 2 on VQ2D.} Here, we evaluate on 100 randomly sampled examples from the VQ2D validation set.}
    \label{tab:sam2-vq2d}
\end{table}

\begin{table}[t]
    \captionsetup{skip=5pt}
    \centering
    \footnotesize
    \begin{tabular}{c c c}
        \toprule
        \textbf{Category} & \textbf{SAM 2} & \vqlb{} \\
        \midrule
        Last occurrence localized & 54 & 61 \\
        Prior occurrence localized & 24 & 32 \\
        Wrong object localized & 18 & 7 \\
        No track returned & 4 & 0 \\
        \bottomrule
    \end{tabular}
    \caption{{\bf Response track prediction analysis of SAM 2 and \vql{}}. We compare the predictions of SAM 2 and \vql{} on 100 sampled examples from the VQ2D validation set. Predictions are categorized into four types, and the count for each category is reported.}
    \label{tab:sam2-analysis}
\end{table}